\documentclass[runningheads]{llncs}

\usepackage{eccv}

\usepackage{eccvabbrv}

\usepackage{graphicx}
\usepackage{booktabs}

\usepackage[accsupp]{axessibility}  

\usepackage{hyperref}

\usepackage{orcidlink}
\usepackage[dvipsnames]{xcolor}
\usepackage{float}
\usepackage{multirow}
\usepackage[most]{tcolorbox}
\newtcolorbox{promptbox}[2][]{  
  colback=blue!5!white,
  colframe=black,
  title=#2,  
  coltitle=white,
  #1  
}

\newcommand{\method}{LAP\xspace}

\definecolor{cvprblue}{rgb}{0.21,0.49,0.74}

\begin{document}

\title{LAP: A Language-Aware Planning Model For Procedure Planning In Instructional Videos} 

\titlerunning{LAP}

\author{Lei Shi\inst{1} \and
Victor Aregbede\inst{1} \and
Andreas Persson\inst{1} \and
Martin Längkvist\inst{1} \and
Amy Loutfi\inst{1} \and
Stephanie Lowry\inst{1}
}

\authorrunning{Shi. et al.}

\institute{Örebro University, SE-701 82 Örebro, Sweden \\
\email{lei.shi@oru.se}\\
}

\maketitle

\begin{abstract}
Procedure planning requires a model to predict a sequence of actions that transform a start visual observation into a goal in instructional videos. While most existing methods rely primarily on visual observations as input, they often struggle with the inherent ambiguity where different actions can appear visually similar. In this work, we argue that language descriptions offer a more distinctive representation in the latent space for procedure planning. We introduce Language-Aware Planning (LAP), a novel method that leverages the expressiveness of language to bridge visual observation and planning.
\method uses a finetuned Vision Language Model (VLM) to translate visual observations into text descriptions and to predict actions and extract text embeddings. These text embeddings are more distinctive than visual embeddings and are used in a diffusion model for planning action sequences.
We evaluate \method on three procedure planning benchmarks: CrossTask, Coin, and NIV. \method achieves new state-of-the-art performance across multiple metrics and time horizons by large margin, demonstrating the significant advantage of language-aware planning.
\end{abstract}
\section{Introduction}
\label{sec:intro}

\begin{figure}
    \centering
    \includegraphics[width=\linewidth]{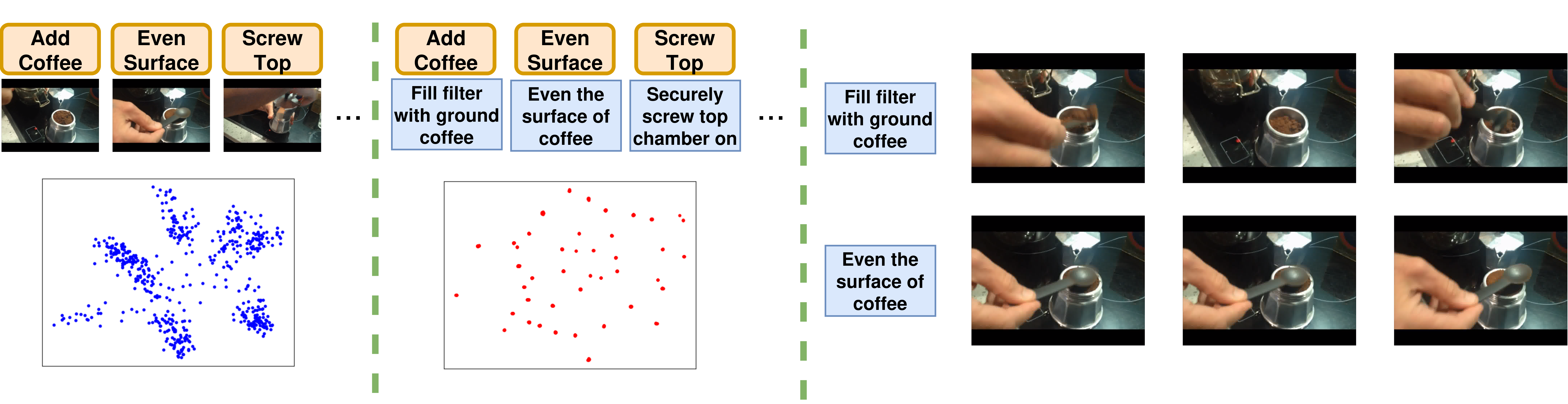}
    \caption{\textcolor{orange}{Orange boxes} show the labels of actions. \textcolor{cvprblue}{Blue boxes} show the language descriptions of actions. \textbf{Left}: Visualisation of extracted features of visual observations in the latent space. Each point represents one video. \textbf{Middle}:  Visualisation of text embeddings of actions in the latent space. \textbf{Right}: Start visual observation of ``Fill filter with ground coffee'' and ``Even the surface of coffee''. 
    }
    \label{fig:teaser}
\end{figure}

A fundamental goal of future human-AI interaction is to enable AI systems that proactively assist humans in accomplishing complex tasks \cite{huang2023grounded, shi2023inferring, patel2023pretrained, kim2024palm}. Procedure planning in instructional videos represents a critical step toward this vision. The procedure planning task \cite{chang2020procedure} is to plan the intermediate action, given start observations and the goal observation. This capability is crucial for building AI systems that can assist in real-world tasks \cite{hu2023look, castelo2023argus, luo2025fmb}.

While visual observations provide rich information about the action step, they have, however, a critical limitation: visual observations of two different actions can sometimes look similar.
The right part of \autoref{fig:teaser} exemplifies this limitation; the upper row shows the start observation of the action ``Add Coffee'', while the lower row shows the start observation of the action ``Even Surface''. Most content in the frames of both actions looks similar (e.g., the background, objects, hand, and so on), even though they are two different actions. 
This makes it difficult for a model to distinguish actions based on visual observation. 
On the other hand, the language descriptions of actions are more distinctive. In the left and middle parts of \autoref{fig:teaser}, we show the visual observations and the language descriptions of actions in the latent space. The visual observations are more cluttered than the language descriptions, which makes it easier for models to plan actions and to ground them in the language modality \cite{persson2019semantic}. 

Most previous works in procedure planning have primarily focused on learning representations from visual observations \cite{chang2020procedure, sun2022plate, li2023skip, wang2023pdpp, nagasinghe2024not}. 
Some of the works leveraged text embeddings of action labels indirectly via visual observations, for instance, as language supervision \cite{zhao2022p3iv}, or as additional noise in diffusion models \cite{shi2024actiondiffusion}. Other works also incorporated Large Language Models (LLMs) to augment descriptions of actions \cite{niu2024schema, yang2025planllm}, or to plan action sequences based only on text \cite{liu2023language}. All of the works either overlook the potential of language descriptions of actions, use text embeddings as auxiliary, or ignore the power of the distinctive representation of language descriptions in the latent space.

To this end, we propose \textbf{\method}: A \underline{\textbf{L}}anguage-\underline{\textbf{A}}ware \underline{\textbf{P}}lanning model for procedure planning in instructional videos. We leverage the expressiveness of natural language descriptions and their distinctiveness in the latent space to bridge the visual observation and action planning domains for the task of procedure planning. 
Specifically, we finetune a pretrained Vision Language Model (VLM) with professor forcing to translate visual observations into text embeddings. We use elaborated descriptions of action as supervision so that different actions will not have similar verb/noun to casuse confusion.
In the planning stage, we plan intermediate actions using diffusion models as they have been shown to be successful with temporal data in various domains \cite{jiao2024diffgaze, yan2023gazemodiff}.
We use the denoising diffusion probabilistic model (DDPM) \cite{ho2020denoising}, which generates an intermediate action sequence as the plan. The model conditions the generation process on the text embeddings of the predicted start and goal actions.
We evaluate our proposed \method on three procedure planning benchmarks: CrossTask \cite{zhukov2019cross}, Coin \cite{tang2019coin}, and NIV \cite{alayrac2016unsupervised}.
The experimental results demonstrate that \method achieves State-Of-The-Art (SOTA) performance on almost all metrics at all time horizons across all three datasets.
We thereby show that text embeddings are more distinctive than visual observations and that procedure planning can further benefit from such embeddings.

Our contributions are threefold. 
First, we introduce \method, a novel procedure planning method that uses the expressive and discriminative properties of language descriptions to bridge visual observations and procedure planning.
Second, we achieve SOTA performance on three challenging procedure planning datasets (CrossTask, Coin, and NIV) across multiple evaluation metrics and planning horizons by large margin.
Third, we empirically demonstrate that procedure planning can benefit from using text embeddings of language descriptions, as they provide a more distinctive representation than visual observations.
\section{Related Work}
\label{sec:related_work}
\subsection{Procedure Planning in Instructional Videos}
Instructional videos provide step-by-step instructions for accomplishing various tasks \cite{miech2019howto100m, zhukov2019cross, zhou2018towards}.
Procedure planning aims to model the temporal structure of a task by predicting a sequence of actions given the start state and the goal state in an instructional video. Early approaches use Recurrent Neural Networks (RNNs) \cite{chang2020procedure} or Reinforcement Learning (RL) \cite{bi2021procedure}, as well as Transformers \cite{sun2022plate}, to model action dependencies.   
E3P \cite{wang2023event} proposes an event-guided paradigm where the action sequences are predicted by visual observation and inferred events.
SkipPlan \cite{li2023skip} decomposed long action sequences into shorter ones and skips less specific actions to improve the sequence prediction. 

\subsection{Large Language Models in Planning}
Another paradigm seeks to incorporate Large Language Models (LLMs) for planning. 
LLMs have been used widely in text-image-based planning \cite{wu2021understanding, lu2023multimodal}, where the task is to generate text descriptions and images for future steps.
In robotic planning tasks, LLMs are used for task decomposition \cite{singh2022progprompt, yang2024selfgoal}, or as a high-level planner \cite{dagan2023dynamic, guan2023leveraging}.
For procedure planning, SCHEMA \cite{niu2024schema} uses LLMs to generate language descriptions of states and injects the description features into the cross-attention in transformers to decode action sequences.
PlanLLM \cite{yang2025planllm} also adopts the use LLM as in SCHEMA and further uses a finetuned VLM to refine action sequences and generate free-form predictions.
\method also uses LLMs and VLMs in the pipeline, however, the motivation of including LLMs/VLMs and how they are used is fundamentally different from SCHEMA and PlanLLM.
We use the LLM to elaborate the action descriptions to avoid same verb/noun in different actions. Essentially we paraphrase the action descriptions to make actions more distinctive to each other and use them as supervision for finetuning the VLM. On the other hand, SCHEMA and PlanLLM use LLMs to augment the states before and after an action to represent common-sense knowledge and use them in cross-attention. 
Moreover, the impact of the use of LLM in SCHEMA is incremental on procedure planning, while ours shows large improvements.
\method uses a VLM to transform data from visual domain to text domain to avoid confusion. PlanLLM uses a VLM to refine predictions of sequences. However, PlanLLM uses visual observation of intermediate actions and ground truth action text of all actions, which gives unfair advantages and makes the contributions of LLM and VLM unclear and unverified.

\subsection{Generative Models for Planning}
Generative models have demonstrated a strong capacity to model the complex distribution of action sequences. P$^3$IV \cite{zhao2022p3iv} adds adversarial generative modules to transformers for training. During inference, the transformers are used as a generator to sample multiple action sequences for planning. 
PDPP \cite{wang2023pdpp} constructs the start and goal observation as well as the predicted task as input and uses Denoising Diffusion Probabilistic Models (DDPMs) to predict the action sequences. ActionDiffusion \cite{shi2024actiondiffusion} further incorporates action embeddings into the noise-adding phase of DDPM to enhance temporal dependencies in actions. KEPP \cite{nagasinghe2024not} explores integrating knowledge graphs learned from full action sequences to guide DDPMs.
CLAD \cite{shi2025clad} uses text embeddings to constrain the latent space of diffusion model to reduce the search space of actions.
MTID \cite{zhoumasked25} introduces a temporal interpolation module to generate interpolated latent features from start and goal observations. The interpolated latent features are further integrated into the denoising process of a diffusion model to improve the temporal relations.

\section{Method}
\label{sec:method}

\begin{figure*}[!hbt]
    \centering
    \includegraphics[width=\textwidth]{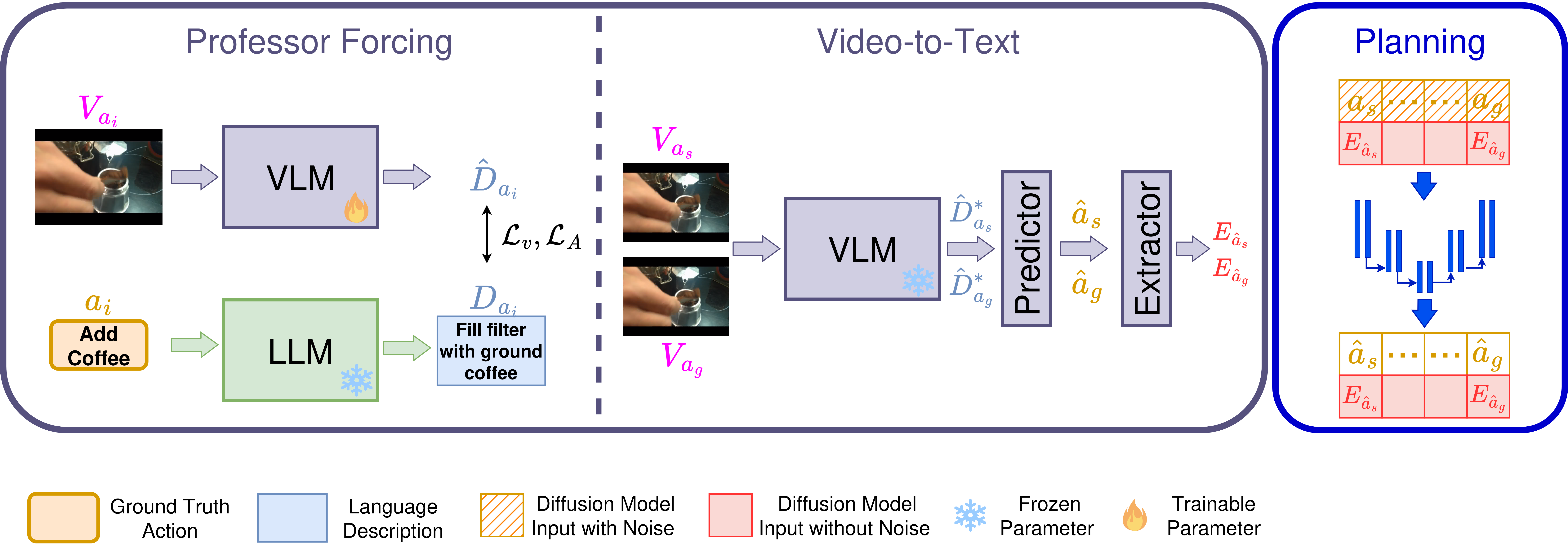}
    \caption{Overview of our proposed \method. \textbf{Professor Forcing}: We finetune a VLM using professor forcing to predict the elaborated description $\hat{D}_{a_i}$ of an action $a_i$. The elaborated description is obtained by prompting $a_i$ to a pretrained LLM. \textbf{Video-to-Text}: 
    we use the finetuned VLM to generate descriptions for start and goal visual observation $V_{a_s}$ and $V_{a_g}$ and find out the descriptions that are most similar to $D_{a_s}$ and $D_{a_g}$ to represent the predicted action $\hat{a}_s$ and $\hat{a}_g$. Then the extractor extracts the corresponding embeddings $E_{\hat{a}_s}$ and $E_{\hat{a}_g}$.
    \textbf{Planning}: We use a diffusion model to plan the action sequence. We use the embeddings of start and end embeddings $E_{\hat{a}_s}$ and $E_{\hat{a}_g}$ as well as action sequence to construct the input to the diffusion model. During training, only the action dimension is noised. } 
    \label{fig:overview}
\end{figure*}

\subsection{Problem Definition and Method Overview}
The task definition of procedure planning \cite{chang2020procedure} is, given the start observation $o_s$ and goal observation $o_g$, a model should generate a plan $\pi$ that contains the intermediate actions $[a_1:a_T]$ with time horizon $T$. 
Our motivation is to use text information instead of visual observations, as text information is potentially more distinctive in the latent space and could benefit the learning of plan $\pi$. This requires transforming visual observation to text information, formally,

\begin{equation}\label{eq:problem_formation}
    p(\pi \mid o_s, o_g) = \int p(\pi \mid \hat{T}_s, \hat{T}_g) \; p(\hat{T}_s, \hat{T}_g \mid o_s, o_g) \, d\hat{T}_sd\hat{T}_g,
\end{equation}

where $\hat{T}_s$ and $\hat{T}_g$ are the text embeddings of predicted start and goal actions. The procedure planning task is essentially split into two steps: (1) obtaining text embeddings of the start and goal visual observations (Section \ref{sec:vid2text}), and (2) planning $\pi$ based on text embeddings $\hat{T}_s$ and $\hat{T}_g$ (Section \ref{sec:planning}).

\autoref{fig:overview} shows the overview of \method. We first finetune a VLM to predict elaborated text descriptions (captions) from visual observations. The elaborated text descriptions are augmented by a pretrained LLM using the original action descriptions. 
Next, we transform the visual observations of the start and goal into text embeddings by incorporating the finetuned VLM. The generate text descriptions are then used for predicting the actions and extracting text embeddings of the start and goal.
Finally, we deploy a diffusion model to generate the plan $\pi$.

\subsection{Professor Forcing}
\label{sec:professor_forcing}
We finetune a VLM to predict text descriptions of an input video $V_{a_i}$ using the elaborated language description $D_{a_i}$ as supervision. The reason we elaborate the language description is that the text of the ground truth action is typically a short phrase (e.g. ``Add Coffee'') and some different actions may have the same verb or noun. When two distinct actions share the same verb or noun, the text embeddings become less distinctive. Finetuning with the elaborated text will make the description of visual observation even more distinctive. To overcome this limitation, we elaborate the action text by incorporating a pretrained LLM \cite{liu2024deepseek}. We query the LLM to obtain detailed language descriptions using the step instructions from WikiHow. As a result, each action $a_i$ has a detailed language description $D_{a_i}$, which complements the original action text. 

\begin{figure}[!ht]
    \centering
    \includegraphics[width=0.7\linewidth]{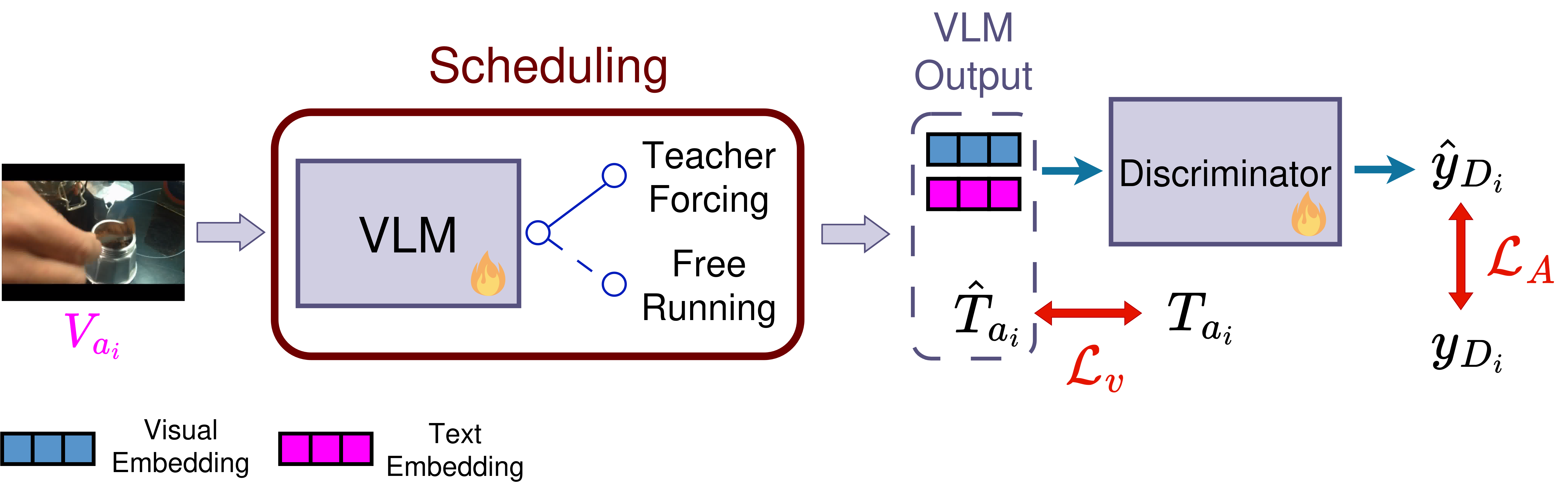}
    \caption{VLM finetuning: we use professor forcing and scheduled sampling to finetune the VLM. During the forward pass, the VLM either use teaching forcing or free running to generate tokens with a scheduled probability. A discriminator is used to minimise the distribution distances between them so that the generated tokens are as close to ground truth tokens.}
    \label{fig:vlm}
\end{figure}

The VLM we use is the video captioning model from \cite{zhao2023learning}.
It is pretrained on Ego4D dataset\cite{grauman2022ego4d}. 
We freeze the visual encoder and the text decoder and only finetune the gated cross-attention modules.
We use professor forcing \cite{lamb2016professor} and scheduled sampling \cite{bengio2015scheduled} for finetuning (\autoref{fig:vlm}). 
Our motivation is to improve performance without increasing computational cost too much. Alternatives for finetuning include using free running and using teacher forcing.
Using free running alone, i.e. free running both at training and inference,
will significantly increase training time due to autoregressive generation. Using teacher forcing, i.e. teaching forcing at training and free running at inference, results in lower performance.
Specifically, given a video input $V_{a_i}$, the VLM outputs tokens $\hat{T}_{a_i}$ through the forward pass. The VLM either use teacher forcing or free running (autoregressive) with a probability $p_t$. In each forward pass, $p_t$ is randomly generated and if it is larger than a ratio, the VLM uses teacher forcing, otherwise it uses free running. The ratio is scheduled with a linear decrease. 
A discriminator then predicts whether the generated text is by teaching forcing or free running. 
We optimise the parameters of VLM and discriminator separately. The discriminator is updated every two iterations, and VLM is updated every iteration with the following loss, 
\begin{align}
    \mathcal{L}_A = BCE(y_D, \hat{y}_D),\\
    \mathcal{L}_v= \mathcal{L}_c + w\mathcal{L}_A,
\end{align}
where $\mathcal{L}_A$ is binary cross-entropy loss used for optimising discriminator and $y_D$ and $\hat{y}_D)$ are the binary ground truth and prediction from discriminator. $\mathcal{L}_v$ is the loss used for optimising trainable parameters in VLM. $\mathcal{L}_c$ is token-level cross entropy loss and $w$ is weight set to 0.1.

\subsection{Video to Text}
\label{sec:vid2text}
After finetuning, we transform the start and goal actions from visual domain to text domain. The VLM generates multiple text descriptions for the start and goal visual observation. We then select the descriptions that can best represent the start and goal action in the video and extract text embeddings for further use in planning.

\noindent \textbf{Predictor:} 
we get the action label of $V_{a_i}$ using $\hat{D}_{a_i}$. 
We use a similar way as in sentence summarization \cite{jia2021flexible} to obtain actions from visual observations. 
After the VLM is finetuned, we generate $M$ numbers of descriptions for each video input $V_{a_i}$. We calculate ROUGE-1 scores for all $M$ generated descriptions against the ground truth descriptions of actions and compare them with a threshold. If a description in $M$ descriptions whose ROUGE-1 score is higher than or equal to the threshold, we consider it represents the action of $V_{a_i}$, otherwise we assign a ``unknown'' label to $V_{a_i}$. When multiple generated descriptions exceed the threshold, the one with the highest score is selected.
The intuition is that the generated descriptions can have the same semantics even though not every word in the description is exactly the same as in the ground truth. We set the threshold to 0.5, representing that if more than half of the words in the generated description exist in the ground truth, we consider that it can represent the visual observation.

\noindent \textbf{Extractor:} once we have the predicted action $\hat{a}_i$, we use the text encoder from \cite{miech2020end}. The text encoder is pretrained on HowTo100M \cite{miech2019howto100m} dataset. For each $\hat{a}_i$, the paired language description $D_{\hat{a}_i}$ is fed to the text encoder to obtain the text embedding $E_{\hat{a}_i}$ of $\hat{a}_i$. 
We also investigate another approach to obtain $E_{\hat{a}_i}$, denoted as \method-Text. From the $M$ generated descriptions, we select the description with the least nll score and extract embeddings directly using the same text encoder as above.

\subsection{Planning}
\label{sec:planning}
We use DDPM \cite{ho2020denoising} diffusion model as the planning model. The input $x_0$ to the diffusion model is,

\begin{equation}
\label{eq:diffusion_input}
x_0 = 
    \begin{bmatrix}
        a_s & ... & ... & ... & a_g \\
        E_{\hat{a}_s} & 0 &... &0 & E_{\hat{a}_g}
    \end{bmatrix},
\end{equation}

where $a_s$ and $a_g$ are the start and goal action, respectively, and $E_{\hat{a}_s}$ and $E_{\hat{a}_g}$ are the text embeddings of the start and goal action. 
Following \cite{wang2023pdpp}, the dimension of text embeddings is not noised. In the forward passing of the diffusion phase, a Gaussian noise $\epsilon \sim \mathcal{N}(0, \mathbf{I})$ is added to the action dimension iteratively. The denoising phase follows,

\begin{equation}
x_n = 
    \begin{bmatrix}
        \epsilon & \epsilon & ... & \epsilon & \epsilon \\
        E_{\hat{a}_s} & 0 &... &0 & E_{\hat{a}_g}
    \end{bmatrix},
\end{equation}

where the text embedding dimension is kept unchanged.
\section{Experiments}
\label{sec:exp}

\subsection{Datasets And Data Curation}\label{sec:datasets}
To evaluate our proposed \method, we used three procedure planning datasets: CrossTask \cite{zhukov2019cross}, Coin \cite{tang2019coin}, and NIV \cite{alayrac2016unsupervised}.
CrossTask dataset has 2,750 videos, 18 tasks, and 105 action classes. The average number of actions in a video is 7.6.
Coin dataset has 11,827 videos with 3.6 actions per video on average, 180 different tasks, and 778 action classes.
NIV dataset is the smallest dataset among all three and has the least number of videos, tasks, and action classes.
It contains 150 videos, five types of tasks, and 18 action classes. The average number of actions in a video is 9.5.

We follow the data curation process used in previous work \cite{chang2020procedure}. 
For each video in the dataset, we extract action sequences using a sliding window with horizon $T$.
Each action in the extracted action sequence $[a_i, ... , a_{i+T-1}]$ corresponds to a video clip.
There are two settings for getting the start visual observation $o_s$ and goal visual observation $o_g$. 
We refer to the two settings as the PDPP setting \cite{wang2023pdpp} and the KEPP setting \cite{nagasinghe2024not}. 
In the PDPP setting, the start observation $o_s$ period is from the beginning of the first action to three seconds after the first action. 
The goal observation $o_g$ begins two seconds before the last action in a sequence and ends one second after the last action. 
In the KEPP setting, $o_s$ begins one second before the first action and ends two seconds after the first action. The goal observation $o_g$ starts at one second before the last action and finishes at two seconds after the last action. 
In this paper, we follow the KEPP setting, as most of the recent baselines also use KEPP setting.

\subsection{Baselines}
We compared our method to four SOTA methods for procedure planning on the instructional video datasets: PDPP \cite{wang2023pdpp}, ActionDiffusion \cite{shi2024actiondiffusion}, SCHEMA \cite{niu2024schema}, SkipPlan \cite{li2023skip}, KEPP \cite{nagasinghe2024not},  PlanLLM \cite{yang2025planllm}, and MTID \cite{zhoumasked25}.
All the baselines use pre-extracted visual features for $o_s$ and $o_g$, and they are extracted by using a model \cite{miech2020end} pre-trained on the HowTo100M dataset \cite{miech2019howto100m}. Our method, on the other hand, needs the original video for training. For the CrossTask dataset, some videos are no longer available from YouTube. Therefore, we remove the unavailable videos from CrossTask for our training. This results in 1870 videos in the new training set and 777 in the new test set, while the original training and test sets have 1921 and 797 videos, respectively.

Since the pre-extracted features used by other baselines are extracted from the original CrossTask dataset, and we can only use a subset of videos, we train all baselines on the new dataset splits to ensure a fair comparison. 
We train all baseline models with the official code provided by the authors, and with data curation using KEPP setting. We also keep all training parameters the same as reported in the original papers.

\subsection{Metrics}
We adopted Success Rate (SR), mean Accuracy (mAcc), and mean Single Intersection over Union (mSIoU) for evaluation.
SR assesses whether a generated action plan is completely successful, i.e. a plan is considered successful only if every action and its sequential order are correct, this makes SR the most challenging metric among all three metrics.
mAcc is computed based on the correct actions in the action plan, without considering their order. Similarly, mSIoU measures Intersection over Union (IoU) between the predicted and ground-truth actions and does not account for action ordering.
Early works \cite{chang2020procedure, zhao2022p3iv, bi2021procedure} computed mIoU across all action plans in a mini-batch. However, \cite{wang2023pdpp} later demonstrated that mIoU depends on the batch size. Consequently, subsequent studies \cite{shi2024actiondiffusion, zhoumasked25} have adopted mSIoU instead.

\section{Results}
\label{sec:results}

\subsection{Comparison with SOTA}

\begin{table}[hbt!]
    \centering
    \caption{Comparisons with baseline methods on CrossTask dataset with time horizon from $T=3$ to $T=6$. $\uparrow$ means higher is better. The best results are shown in \textbf{bold}. The second best numbers are shown with \underline{underline}. All baselines and our method use KEPP setting for data curation.}
    \label{tab:crosstask_sota_professor}
    \resizebox{\linewidth}{!}{
    \begin{tabular}{lcccccccc}
        \toprule
        & \multicolumn{3}{c}{T=3} & \multicolumn{3}{c}{T=4} & T=5 & T=6 \\
        \cmidrule(lr){2-9} 
        & SR$\uparrow$ & mAcc$\uparrow$ & mSIoU$\uparrow$ & SR$\uparrow$ & mAcc$\uparrow$ & mSIoU$\uparrow$ & SR$\uparrow$ & SR$\uparrow$ \\
        \midrule
        PDPP \cite{wang2023pdpp} (CVPR 2023) & 33.9 & 61.39 & / & 20.77 & 56.57 & / & 12.72 & 8.07 \\
        SkipPlan \cite{li2023skip} (ICCV 2023) & 27.6 & 59.9 & / & 14.06 & 54.64 & / & 6.51 & 5.24 \\
        ActionDiffusion \cite{shi2024actiondiffusion} (WACV 2025) & 33.25 & 61.47 & \underline{65.09} & 20.01 & 56.74 & \textbf{63.98} & 12.48 & 8.07 \\
        KEPP \cite{nagasinghe2024not} (CVPR 2024) & 33.27 & 60.16 & / & 19.41 & 54.45 & / & 12.01 & 7.89 \\
        SCHEMA \cite{niu2024schema} (ICLR 2024) & 32.34 & 57.3 & / & \underline{22.51} & 51.47 & / & \underline{12.89} & 7.54 \\
        \method (Ours) & \textbf{41.14} & \textbf{70.13} & \textbf{65.98} & \textbf{23.91} & \textbf{63.70} & \underline{62.56} & \textbf{14.69} & \textbf{8.79} \\
        \method-Text (Ours) & \underline{35.84} & \underline{65.65} & 62.48 & 21.14 & \underline{59.64} & 59.99 & 12.23 & \underline{8.66} \\
        \bottomrule
    \end{tabular}
    }
\end{table}

\noindent \textbf{CrossTask}:
\autoref{tab:crosstask_sota_professor} shows the comparison with SOTA baselines on the CrossTask dataset across time horizons from T=3 to T=6. 
Our method demonstrates strong performance across all time horizons. \method achieves the highest SR across all horizons. 
At $T=3$ and $T=4$, \method also has the best mAcc scores. 
\method-Text also has competitive performance against baselines. For $T=3$, \method-Text outperforms all baselines on SR and mAcc. On longer horizons $T=4$ and $T=5$ , \method-Text achieves comparable performance against baselines, but higher SR when time horizon is six.

\begin{table*}[h]
    \centering
    \caption{Comparisons with baseline methods on Coin dataset with time horizon $T=3$ and $T=4$. $\uparrow$ means higher is better. The best results are shown in \textbf{bold}. The second best numbers are shown with \underline{underline}. All baselines and our method use KEPP setting for data curation. $\dagger$ indicates the model uses intermediate visual observations.}
    \label{tab:coin_sota_professor}
    \begin{tabular}{lcccccc}
        \toprule
        & \multicolumn{3}{c}{T=3} & \multicolumn{3}{c}{T=4} \\
        \cmidrule(lr){2-7} 
        & SR↑ & mAcc ↑ & mSIoU↑ & SR↑ & mAcc ↑ & mSIoU↑ \\
        \midrule
        PDPP \cite{wang2023pdpp} (CVPR 2023) & 21.33 & 45.62 & 51.82 & 13.67 & 42.58 & / \\
        SkipPlan \cite{li2023skip} (ICCV 2023) & 23.65 & 47.12 & / & 16.04 & 43.19 & / \\
        KEPP \cite{nagasinghe2024not} (CVPR 2024) & 20.25 & 39.87 & / & 15.63 & 39.53 & / \\
        SCHEMA \cite{niu2024schema} (ICLR 2024) & 32.09 & 49.84 & / & 22.02 & 45.33 & / \\
        PlanLLM $\dagger$ \cite{yang2025planllm} (AAAI 2025) & \underline{33.22} & \underline{54.33} & / & \underline{25.31} & 48.79 & / \\
        MTID \cite{zhoumasked25} (ICLR 2025) & 30.44 & 51.70 & \underline{59.74} & 22.74 & \underline{49.9} & \underline{61.25} \\
        \method (Ours) &\textbf{44.43} & \textbf{68.08} & \textbf{66.71} & \textbf{31.56} & \textbf{63.77} & \textbf{68.24} \\
        \method-Text (Ours) & 27.06 & 50.06 & 52.43 & 19.25 & 46.72 & 55.25 \\
        \bottomrule
    \end{tabular}
\end{table*}

\noindent \textbf{Coin}:
\autoref{tab:coin_sota_professor} shows the comparisons of our proposed \method against SOTA baselines on the Coin dataset with time horizons $T=3$ and $T=4$.
The results show that \method achieves the best performance on all metrics. For $T=3$, \method has the highest SR (44.43), outperforming the best baseline PlanLLM, whose SR is 33.22. For $T=4$, \method also achieves the best SR, surpassing PlanLLM by 6.25. 
Note that even though PlanLLM has the advantage of using intermediate visual observations, our \method still outperforms it by large margin. 
The \method-Text generally shows lower performance than \method across most metrics. This is probably because of the large action space of the dataset, which lowers the matching words in the generated caption. Directly using the text description causes more variations in the test set for the planning model.

\begin{table*}[!hbt]
    \centering
    \caption{Comparisons with baseline methods on NIV dataset with time horizon $T=3$ and $T=4$. $\uparrow$ means higher is better. The best results are shown in \textbf{bold}. The second best numbers are shown with \underline{underline}. All baselines and our method use KEPP setting for data curation. $\dagger$ indicates the model uses intermediate visual observations.}
    \label{tab:niv_sota_professor}
    \begin{tabular}{lcccccc}
        \toprule
        & \multicolumn{3}{c}{T=3} & \multicolumn{3}{c}{T=4} \\
        \cmidrule(lr){2-7} 
        & SR$\uparrow$ & mAcc $\uparrow$ & mSIoU$\uparrow$ & SR$\uparrow$ & mAcc $\uparrow$ & mSIoU$\uparrow$ \\
        \midrule
        PDPP \cite{wang2023pdpp} (CVPR 2023) & 22.22 & 39.50 & / & 21.30 & 39.24 & / \\
        KEPP \cite{nagasinghe2024not} (CVPR 2024) & 24.44 & 43.46 & / & 22.71 & 41.59 & / \\
        SCHEMA \cite{niu2024schema} (ICLR 2024) & 27.93 & 41.64 & / & 23.26 & 39.93 & / \\
        PlanLLM $\dagger$ \cite{yang2025planllm} (AAAI 2025) & 26.74 & 42.97 & / & 27.08 & 46.96 & / \\
        MTID \cite{zhoumasked25} (ICLR 2025) & 28.52 & 44.44 & 56.46 & 24.89 & 44.54 & 57.46 \\
        \method (Ours) & \textbf{56.51} & \textbf{72.86} & \textbf{72.22} & \textbf{43.81} & \textbf{68.03} & \textbf{70.17} \\
        \method-Text (Ours) & \underline{43.87} & \underline{65.55} & \underline{66.16} & \underline{34.07} & \underline{61.84} & \underline{66.49} \\
        \bottomrule
    \end{tabular}
\end{table*}

\noindent \textbf{NIV}:
\autoref{tab:niv_sota_professor} shows the results on NIV dataset with time horizon $T=3$ and $T=4$, All methods are evaluated under the KEPP data curation setting.
Our method demonstrates a substantial margin of performance increase compared to all baselines with both of its variants.
For $T=3$, \method achieves the best performance, with an SR of 56.51, mAcc of 72.86. For time horizon $T=4$, \method has the highest SR and mAcc, surpassing the best baseline by 18.92 and 23.49.
\method-Text has lower performance in SR and mAcc compared to \method. But it still outperforms the strongest baseline (MTID) by large margins.

\subsection{Ablation Studies}
\subsubsection{Language Enhancement}
We show the impact of language enhancement on predicting the start and goal actions (\autoref{tab:act_pred}). We do not perform experiments on Coin as the descriptions are already in detail. \method achieves higher accuracies than \method without language enhancement, demonstrating the importance of language enhancement. 
\begin{table}[h]
    \centering
    \caption{Action prediction ablations on language enhancement.}
    \label{tab:act_pred}
    \begin{tabular}{lcccccccc}
        \toprule
        & \multicolumn{4}{c}{CrossTask} & \multicolumn{4}{c}{NIV} \\
        \cmidrule(lr){2-9} 
        & \multicolumn{2}{c}{T=3} & \multicolumn{2}{c}{T=4} & \multicolumn{2}{c}{T=3} & \multicolumn{2}{c}{T=4} \\
        \cmidrule(lr){2-9} 
        & start & goal & start & goal & start & goal & start & goal \\
        \midrule
        \method w/o le & 63.89	 & 67.61 & 64 & 68.3 & 55.02 & 52.42 & 58.85 & 55.44 \\

        \method        & \textbf{75.72} & \textbf{79.69} & \textbf{75.77} & \textbf{80.47} & \textbf{67.29} & \textbf{70.63} & \textbf{66.81} & \textbf{68.58} \\

        \bottomrule
    \end{tabular}
\end{table}

\subsubsection{Professor Forcing}
\begin{table}[h]
    \centering
    \caption{Ablations on professor forcing and using text or visual observation for procedure planning on all datasets with $T=3$ and $T=4$. \method-tf indicates training with teaching forcing. \method-vo indicates using visual feature.
    The best numbers are in \textbf{bold}.
    }
    \label{tab:professor_forcing}
    \begin{tabular}{llcccccc}
        \toprule
        &       & \multicolumn{3}{c}{T=3} & \multicolumn{3}{c}{T=4} \\
        \cmidrule{3-8}
         &  & SR$\uparrow$ & mAcc$\uparrow$ & mSIoU$\uparrow$ & SR$\uparrow$ & mAcc$\uparrow$ & mSIoU$\uparrow$ \\
        \cmidrule{1-8}
        \multirow{3}{*}{CrossTask} 
            &\method-vo &32.03 &60.16 &63.34 &19.78 &55.81 &\textbf{63.55} \\
            & \method-tf & 37.82	&68.23	&64.52	&21.57	&60.85	&59.45 \\
            & \method & \textbf{41.14} & \textbf{70.13} & \textbf{65.98} & \textbf{23.91} & \textbf{63.70} & 62.56 \\            
        \cmidrule{1-8}
        \multirow{3}{*}{Coin} 
            & \method-vo &17.54 &38.39 &44.83  &13.2 &36.88 &45.03 \\
            & \method-tf &43.11	&66.49	&65.88	&28.9	&60.55	&65.18 \\
            & \method &\textbf{44.43} & \textbf{68.08} & \textbf{66.71} & \textbf{31.56} & \textbf{63.77} & \textbf{68.24} \\            
        \cmidrule{1-8}
        \multirow{3}{*}{NIV} 
            & \method-vo &27.04 &43.09 &56.96 &20.96 &42.9 &57.78 \\
            & \method-tf &49.81	&68.53	&68.37	&39.82	&67.26	&68.16 \\
            & \method & \textbf{56.51} & \textbf{72.86} & \textbf{72.22} & \textbf{43.81} & \textbf{68.03} & \textbf{70.17} \\         
        \hline
    \end{tabular}
\end{table}
To demonstrate the advantage of professor forcing, we compare \method trained with professor forcing and teacher forcing. In the teacher forcing setting, the VLM is trained with teaching forcing in training and free running during inference. 
\autoref{tab:professor_forcing} shows the results of comparisons on all datasets for $T=3$ and $T=4$. \method trained with professor forcing (\method) consistently outperforms teacher forcing (\method-tf) across all datasets on all metrics, showcasing the importance of incorporating professor forcing. 

\subsubsection{Plan with Text vs Plan with Visual Observation}

To evaluate the contribution of language-aware procedure planning, we conduct an ablation study comparing language-aware planning with visual observation. We compare \method with text embeddings and \method with visual observations (only). For planning based solely on visual observations (denoted as \method-vo), we use the features extracted from the start and goal observations in the diffusion model. Concretely, \autoref{eq:diffusion_input} becomes,

\begin{equation}
x_0 = 
    \begin{bmatrix}
        a_s & ... & ... & ... & a_g \\
        E_{V_s} & 0 &... &0 & E_{V_g}
    \end{bmatrix},
\end{equation}
where $E_{V_s}$ and $E_{V_g}$ are the features of the start visual observation and the goal visual observation, respectively. Following previous work \cite{wang2023pdpp, shi2024actiondiffusion}, $E_{V_s}$ and $E_{V_g}$ are extracted by using the visual encoder from \cite{miech2020end}. 

\autoref{tab:professor_forcing} shows the results on CrossTask, Coin and NIV dataset with time horizon $T=3$ and $T=4$. The results demonstrate that \method with text descriptions in the diffusion model consistently and significantly outperforms \method using only the visual observation (\method-vo) across all metrics and time horizons on the Coin and NIV datasets. 
For the CrossTask dataset, the performance gain of \method on SR is smaller compared to the performances on Coin and NIV datasets in both time horizons.
The significant performance improvement on the Coin and NIV datasets implies that text information provides stronger, more distinctive features, which benefit procedure planning more than visual observations.
In contrast, on the CrossTask dataset, the improvement from using text is not as substantial; this is probably because the visual observation feature is already distinctive enough.

\begin{figure*}[ht]
    \centering
    \begin{subfigure}{0.26\textwidth}
        \centering
        \includegraphics[width=\linewidth]{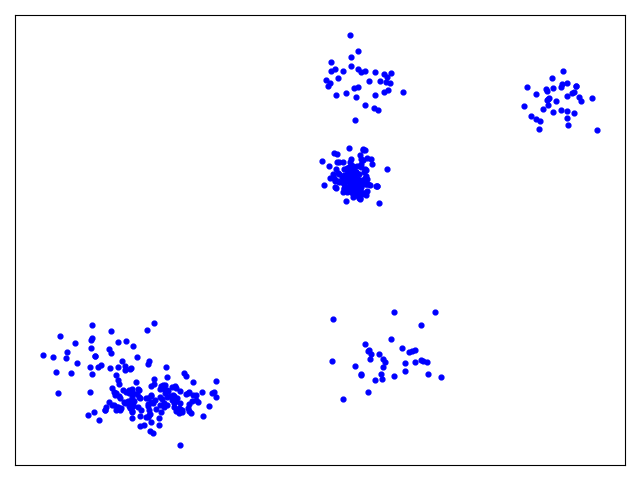}
    \end{subfigure}
    \begin{subfigure}{0.26\textwidth}
        \centering
        \includegraphics[width=\linewidth]{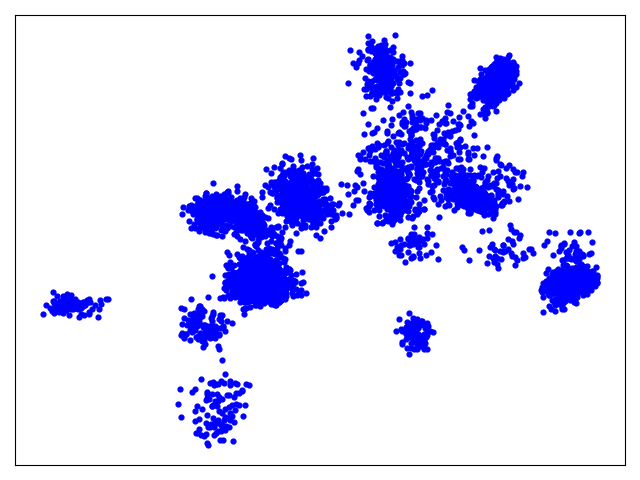}
    \end{subfigure}
    \begin{subfigure}{0.26\textwidth}
        \centering
        \includegraphics[width=\linewidth]{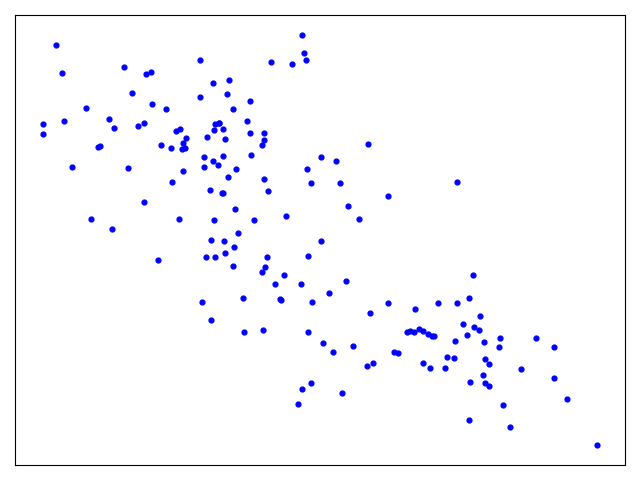}
    \end{subfigure}
    
    
    \begin{subfigure}{0.26\textwidth}
        \centering
        \includegraphics[width=\linewidth]{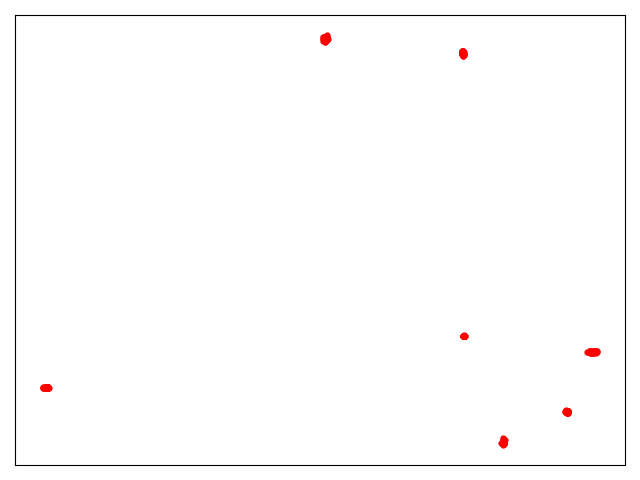}
        \caption{CrossTask}
    \end{subfigure}
    \begin{subfigure}{0.26\textwidth}
        \centering
        \includegraphics[width=\linewidth]{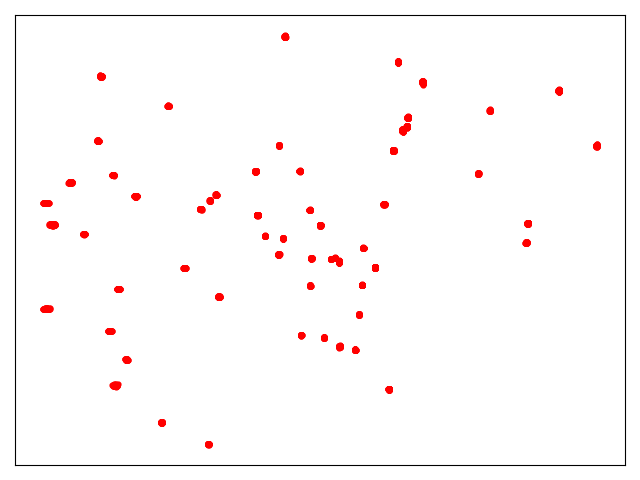}
        \caption{Coin}
    \end{subfigure}
    \begin{subfigure}{0.26\textwidth}
        \centering
        \includegraphics[width=\linewidth]{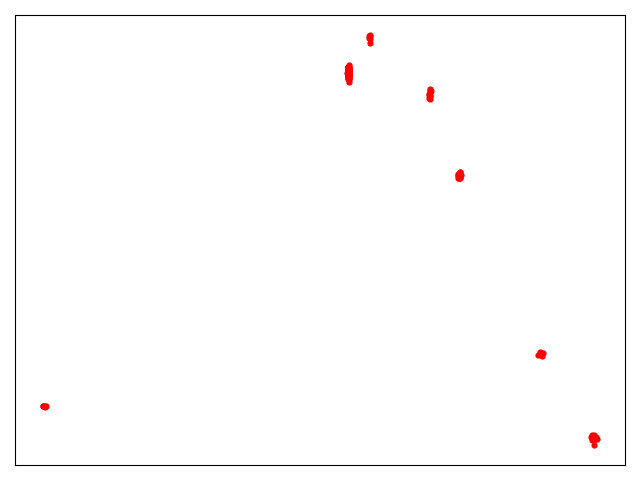}
        \caption{NIV}
    \end{subfigure}
    
    \caption{Visualisation of features of visual observation and text embedding in latent space on CrossTask, Coin and NIV dataset. The first row shows the visual observation and the second row shows the text embedding. All visualisations are the samples from test sets with time horizon $T=3$.}
    \label{fig:vis_latent}
\end{figure*}
To verify this, we present visual observations and text in the latent space.
\autoref{fig:vis_latent} shows the features of visual observation and text embeddings in the latent space for all three datasets. The first row indicates the visual observation (shown in blue) and the second row shows the text embedding (shown in red). All visualisations use the test sets of the respective datasets.
The text embeddings for all datasets are clearly separable in the latent space. The number of latent codes of text embeddings are not less than the number of latent codes of visual observations. They are more clustered and appear as a single point.   
For the visual observation, CrossTask already forms clear and separable clusters in the latent space than Coin and NIV. This confirms a possible reason why using text does not improve CrossTask as much as on Coin and NIV.

\subsubsection{Video to Text}
\begin{table}[!hbt]
    \centering
    \caption{Ablation study of different VLMs on NIV dataset. We compare the VLM used in \method and LLaVa-NeXT-Video 7B. \textbf{Bold} numbers indicate the best results.}
    \label{tab:narrator_vlm_professor}
    \begin{tabular}{ccccccc}
        \toprule
        &  T=3 & & & T=4  & & \\
        \cmidrule{2-4} \cmidrule{5-7}
         & SR$\uparrow$ & mAcc $\uparrow$ & mSIoU $\uparrow$ & SR$\uparrow$ & mAcc $\uparrow$ & mSIoU $\uparrow$ \\
         
        \midrule
        LLaVa-NeXT-Video & 21.9	& 36.68	&37.95 &13.27	&36.28	&37.56  \\
        LLaVa-NeXT-Video-Text & 17.84	&41.76	&46.64 & 13.72	&41.15	&47.89 \\
        \method & \textbf{56.51} & \textbf{72.86} & \textbf{72.22} & \textbf{43.81} & \textbf{68.03} & \textbf{70.17} \\         
        \method-Text & 43.87 & 66.55 & 66.16 & 34.07 & 61.84 & 66.49 \\
        \bottomrule
    \end{tabular}
\end{table}

\method uses the VLM from \cite{zhao2023learning} to transform visual observations of the start and goal into text. The VLM is pretrained for video captioning. In this section, we compare the VLM in \method with the multi-purpose VLM, i.e., LLaVa-NeXT-Video \cite{zhang2024llavanextvideo}. 
We use LoRA \cite{hu2022lora} to finetune LLaVa-NeXT-Video 7B to compare with the VLM used in our \method.
\autoref{tab:narrator_vlm_professor} shows the comparison of \method VLM and LLaVa-NeXT-Video on NIV dataset with time horizon $T=3$ and $T=4$.
The VLM in \method significantly and consistently outperforms LLaVa-NeXT-Video across all metrics and time horizons.
LLaVa-NeXT-Video-Text performs poorly across all metrics, only achieving 17.84 in SR when $T=3$ and 13.72 when $T=4$. The SR of LLaVa-NeXT-Video decreases from 21.9 to 13.27 as the time horizon increases from three to four. 

The reasons why LLaVa-NeXT-Video struggles are likely due to the tasks and video lengths in the datasets used for its training.
The VLM of \method is pretrained on action recognition datasets whose videos are short video clips that contain specific actions. 
On the other hand, LLaVa-NeXT-Video focuses on multiple datasets with video lengths that are typically much longer. 
The video lengths in procedure planning datasets are also short clips, which are more aligned with the datasets used for (pre)training the VLM of \method. Moreover, LLaVa-NeXT-Video is optimised for multiple tasks \cite{fu2025video, yu2019activitynet, xiao2021next}, not targeting the single task of captioning, which is more relevant for procedure planning.

\subsection{Qualitative Results}

\begin{figure*}[!hbt]
    \centering
    \includegraphics[width=0.8\linewidth]{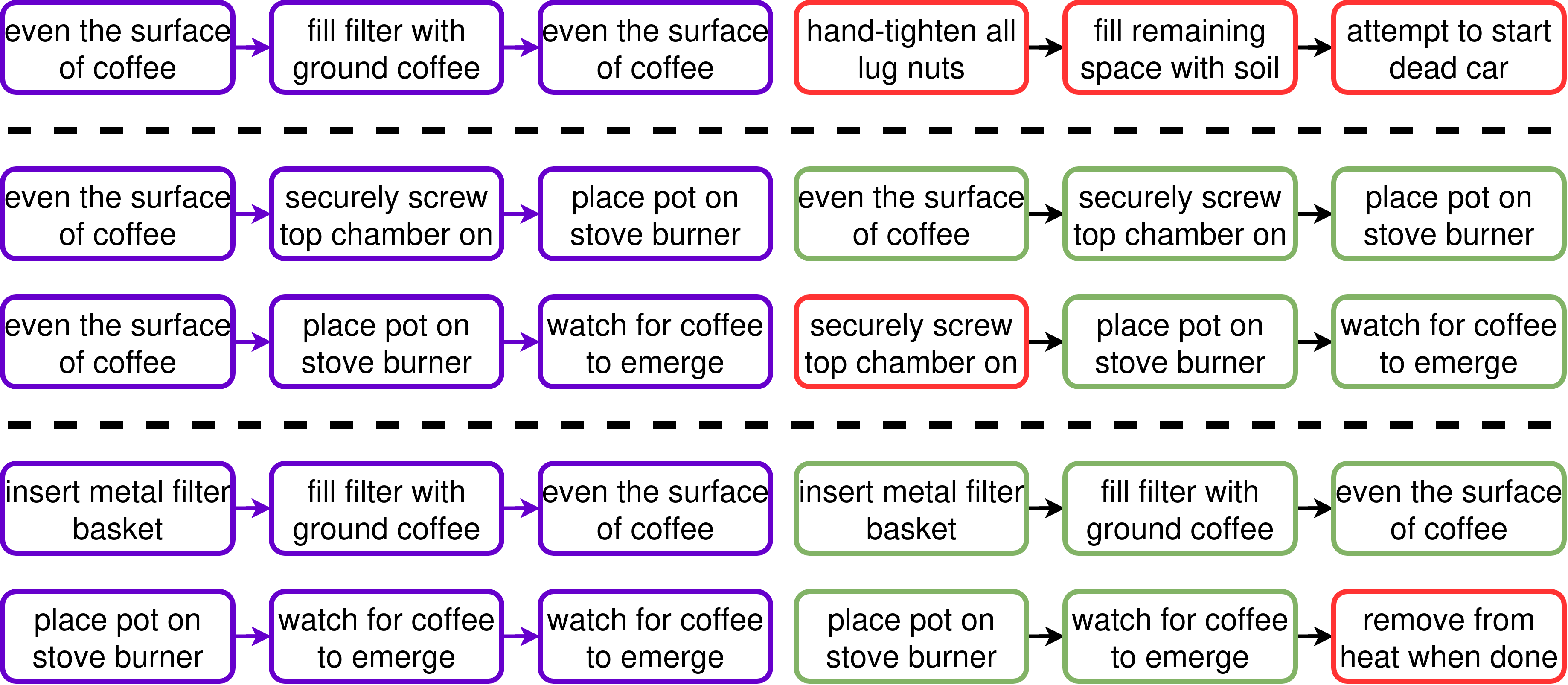}
    \caption{Qualitative results on NIV dataset. Purple boxes indicate the ground truth actions. Green boxes indicate the correctly predicted actions in action sequences. Red boxes indicate the wrong predictions. \textbf{Top:} predictions when $\hat{a}_s$ and $\hat{a}_g$ are ``unknown'' from VLM. \textbf{Middle:} predictions when $\hat{a}_s$ is ``unknown'' and $\hat{a}_g$ is correct. \textbf{Bottom:} predictions when $\hat{a}_s$ is correct and $\hat{a}_g$ is ``unknown''.}
    \label{fig:qualitative}
\end{figure*}

\autoref{fig:qualitative} shows the qualitative results on NIV dataset with time horizon $T=3$. We show three cases of $\hat{a}_s$ and $\hat{a}_g$ from VLM and used for the planning model. First, when $\hat{a}_s$ and $\hat{a}_g$ are ``unknown'' from VLM. As shown in the top part of \autoref{fig:qualitative}, all actions in the predicted sequence are wrong and they are not relevant to the task.
The second (middle part in \autoref{fig:qualitative}) and third (bottom part in \autoref{fig:qualitative}) case are when $\hat{a}_s$ is ``unknown'' and $\hat{a}_g$ is correct and when $\hat{a}_s$ is correct and $\hat{a}_g$ is ``unknown''. When either the start and goal action prediction is correct, the planning model is still able to generate the correct action sequence. In the failure cases, some generated actions in a sequence are correct and the wrong actions are still relevant to the task. 
The reason is that the unknown label is not seen in training, and when only one of the start and goal is unknown, the other correct start/goal can overrule the ``unknown'' feature and still guide the diffusion model to generated correct actions. 
When both predicted start and goal actions are wrong, there will be no guidance and the generation process is random. When all data samples with incorrect start and goal, all of them have wrong generated action sequence. For data samples when the start action is correct, 37.7\% of them has correct generated sequence. 
For the samples when the goal action is correct, 37.5\% of them has correct generated sequence. 

\section{Conclusion}
\label{sec:conclusion}
In this work, we introduce \method, a Language-Aware Planning model for procedure planning in instructional videos. Our approach addresses the inherent ambiguity in visual observations by leveraging the expressiveness and distinctiveness of language descriptions in the latent space. 
We conduct experiments on procedure planning benchmarks: CrossTask, Coin and NIV. The experimental results show that \method achieves SOTA performance on multiple metrics by a large margin.
Ablation studies confirm the importance of using language features for procedure planning due to their distinctive representation, especially in cases where visual observations possess more ambiguities.




%
%
\bibliographystyle{splncs04}
\bibliography{ref}

\newpage
\appendix
\section*{Supplementary}
\section{Implementation Details}
\subsection{Language Augmentation}
The text of the ground truth action is typically a short phrase, e.g. ``Add Coffee''. However, there is a limitation in this form. When two distinct actions share the same verb or noun, they become less distinctive. For example, in a task to make Latte in CrossTask dataset, three out of six actions contain the verb ``pour'' in the text.
To overcome this limitation, we elaborate the action text by incorporating a pretrained LLM \cite{liu2024deepseek}. 
According to the task the actions belong to, we query the LLM to obtain detailed language descriptions using the step instructions from WikiHow. As a result, each action $a_i$ has a detailed language description $D_{a_i}$, which complements the original action text.
\begin{figure}[h]
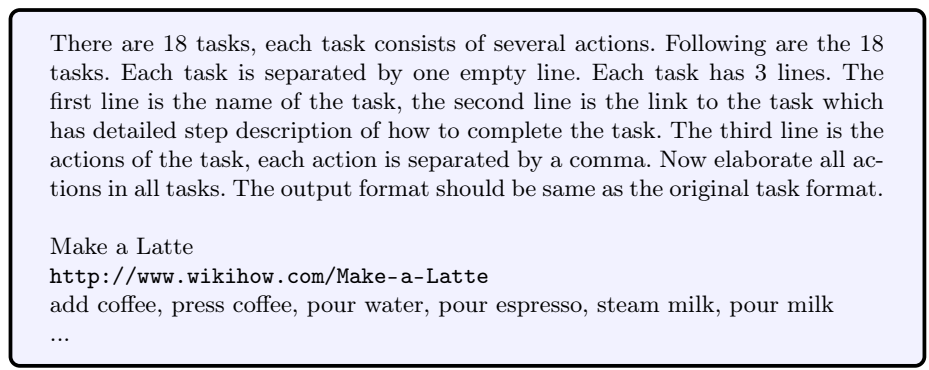

    \centering
    \begin{promptbox}{}
There are 18 tasks, each task consists of several actions. Following are
the 18 tasks. Each task is separated by one empty line. Each task has 3
lines. The first line is the name of the task, the second line is the link to
the task which has detailed step description of how to complete the task.
The third line is the actions of the task, each action is separated by a
comma. Now elaborate all actions in all tasks. The output format should
be same as the original task format.
\\

Make a Latte

\nolinkurl{http://www.wikihow.com/Make-a-Latte}

add coffee, press coffee, pour water, pour espresso, steam milk, pour milk

...
    \end{promptbox}
    \caption{An example of the prompt we use to obtain language descriptions of actions.}
    \label{fig:prompt}
\end{figure}

\subsection{VLM Professor Forcing Finetuning}

We set the initial ratio of teaching forcing to 0.8 and use linear scheduling. The ratio decreases to 0.1 as the training step increases. The total training step equals the length of dataset multiplies by training epochs. 
The discriminator has a text encoder, a vision encoder, a multi-head cross-attention and a classification MLP. The text encoder and the vision encoder are two LSTMs with hidden size 64. The hidden size of multi-head cross-attention is 128 and number of heads is eight. 
We train the discriminator every two iterations. The optimizer is AdamW \cite{loshchilov2017decoupled}. 
The learning rate is $4e^{-5}$ and the batch size is nine for all datasets. For NIV, we train 10 epochs and for CrossTask and Coin we train three epochs. During inference we generate 20 sentences for each input.

\subsection{Diffusion Model}

We use AdamW as the optimizer and set the hidden size of UNet to 256 for all datasets, with a consistent batch size of 128. The training configurations are as follows: for Crosstask, we train for 120 epochs (200 steps/epoch) with 200 diffusion steps; for Coin, 800 epochs (200 steps/epoch) with 200 diffusion steps; and for NIV, 130 epochs (50 steps/epoch) with 50 diffusion steps. All models employ a linear learning rate warm-up for the first 20 epochs, increasing to $5e^{-4}$ for Crosstask and $1e^{-4}$ for Coin. For NIV, the learning rate warms up to $3e^{-4}$ over the first 90 epochs. A learning rate decay is applied to Crosstask (decay by 0.5 every 5 epochs for the last 30 epochs) and Coin (decay by 0.5 over the last 50 epochs).

\section{Additional Ablation Studies}

\subsection{Language Enhancement}
\begin{table}[ht]
    \centering
    \caption{Ablation study of language enhancement on NIV dataset. We train the diffusion model of \method directly with or without language enhancement.  $\dagger$ shows the results with language enhancement. \textbf{Bold} numbers indicate the best numbers.}
    \label{tab:launguage_enhance_professor}
    \begin{tabular}{llccc}
        \toprule
        & &SR & mAcc $\uparrow$ & mSIoU $\uparrow$ \\
        \midrule
        T=3 & \method & 41.64 & 55.14 & 54.61 \\        					
           & \method-Text & 11.52 & 28.38 & 37.52 \\
           & \method $\dagger$& \textbf{56.51}	&\textbf{72.86} &\textbf{72.22} \\
           & \method-Text $\dagger$ & 43.87 & 65.55 & \textbf{66.16} \\
        \midrule
        T=4 & \method & 30.1 & 52.1 & 54.53 \\
           & \method-Text & 7.08 & 27.32 & 38.95 \\
           & \method $\dagger$ & \textbf{43.81} & \textbf{68.03} & \textbf{70.17} \\ 		
           & \method-Text $\dagger$ & 34.07 & 61.84 & \textbf{66.49} \\
        \bottomrule
    \end{tabular}
\end{table}

To evaluate the impact of the enhanced language descriptions, we perform ablation studies on training the planning model with and without language enhancement. We train the planning model directly with the text generated by the VLM. 
\autoref{tab:launguage_enhance_professor} shows the results on the NIV dataset. For both time horizons, \method trained with enhanced language outperforms the model trained with the original text across all metrics.
The consistent improvements indicate the effectiveness of integrating language enhancement into procedure planning.
Training with the original text, \method-Text has significantly worse SR, with 11.52 for $T=3$ and 7.08 for $T=4$. This shows the generated text from videos is less similar to the ground truth text. 
\method trained with original text has much better performance on SR than \method-Text. This is due to the thresholding on the generated text. Mostly the original text consists of a verb and noun. Setting the threshold to 0.5 will consider the generated text is correct even if only the verb or noun is correct. 

\begin{table}[ht]
    \centering
    \caption{Evaluation of generated text on NIV dataset. $\dagger$ indicates the VLM of \method is trained with enhanced language description. Best numbers are shown in \textbf{bold}.}
    \label{tab:text_eval_professor}
    \begin{tabular}{llcc}
        \toprule
        & & ROUGE-1 & ROUGE-2 \\
        \midrule
        T=3 & \method VLM & 0.16 & 0.01 \\
        & \method VLM $\dagger$ & \textbf{0.71} & \textbf{0.68} \\
        \midrule
        T=4 & \method VLM & 0.17 & 0.01 \\
        & \method VLM $\dagger$ & \textbf{0.7}& \textbf{0.68} \\
        \bottomrule
    \end{tabular}
\end{table}
To verify this, we show the 
ROUGE-1 and ROUGE-2 scores of \method VLM trained with original text and with enhanced language description in \autoref{tab:text_eval_professor} . 
The VLM trained with enhanced language description has consistent performance for both time horizons on both metrics. 
The VLM trained with original text, on the other hand, generates less accurate descriptions. 
ROUGE-1 is 0.16 and 0.17 for $T=3$ and $T=4$. ROUGE-2 scores are both 0.01 for both horizons. 
The performance in the quality of generated text is consistent with the performance on procedure planning in \autoref{tab:text_eval_professor}
.

\subsection{Comparison of VLM}

\begin{table*}[h]
    \centering
    \caption{Comparison with  LLaVA-Next-Video 7B on CrossTask and Coin dataset with time horizon $T=3$ and $T=4$. Best numbers are shown in \textbf{bold}.}
    \label{tab:ablation_vlm_professor_crosstask_coin}
    \begin{tabular}{llcccccc}
        \toprule
        \multirow{2}{*}{} & \multirow{2}{*}{} & \multicolumn{3}{c}{T=3} & \multicolumn{3}{c}{T=4} \\
        \cmidrule(lr){3-8} 
        & & SR↑ & mAcc ↑ & mSIoU↑ & SR↑ & mAcc ↑ & mSIoU↑ \\
        \midrule
        \multirow{4}{*}{CrossTask} & LLaVa-NeXT-Video & 17.72	&46.92	&43.48	&9.73	&43.19	&40.64 \\
        & LLaVa-NeXT-Video-Text & 17.15 & 47.09 & 46.73 & 10.08 & 44.32 & 46.8 \\
        & \method & \textbf{41.14} & \textbf{70.13} & \textbf{65.98} & \textbf{23.91} & \textbf{63.7} & \textbf{62.56} \\
        & \method-Text & 35.84 & 65.65 & 62.48 & 21.14 & 59.64 & 59.99 \\
        \midrule
        \multirow{4}{*}{Coin} & LLaVa-NeXT-Video & 29.6 &46.54	&46.07	&20.95	&43.39	&47.08 \\
        & LLaVa-NeXT-Video-Text & 18.33 & 36.76 & 41.27 & 14.52 & 36.58 & 46.93 \\
        & \method & \textbf{44.43} & \textbf{68.08} & \textbf{66.71} & \textbf{31.56} & \textbf{63.77} & \textbf{68.24} \\ 		
        & \method-Text & 27.06 & 50.06 & 52.43 & 19.25 & 46.72 & 55.25 \\
        \bottomrule
    \end{tabular}
\end{table*}
In addition to LLaVA-Next-Video 7B \cite{zhang2024llavanextvideo} on NIV dataset, we show comparisons on CrossTask and Coin dataset in \autoref{tab:ablation_vlm_professor_crosstask_coin}.
LLaVA-Next-Video has low SR, indicating limited capability for procedure planning. On CrossTask, SR drops from 17.72 for $T=3$ to 9.73 for $T=4$. On COIN, SR falls from 29.6 to 20.95 when time horizon increases from three to four. 
LLaVA-Next-Video-Text achieves comparable SR with LLaVA-Next-Video on CrossTask, and even lower SR on Coin. 
The performance gap between LLaVA-Next-Video and our proposed \method is substantial on both datasets. This is consistent with the comparison on NIV dataset shown in the main paper. The significant margin underscores the inability of LLaVA-Next-Video for procedure planning.

\subsection{Effect of Threshold}

\begin{figure*}[h]
    \centering
    \captionsetup[subfigure]{skip=-8pt, font=scriptsize}
    
    \begin{subfigure}{0.32\textwidth}
        \centering
        \includegraphics[width=0.95\linewidth]{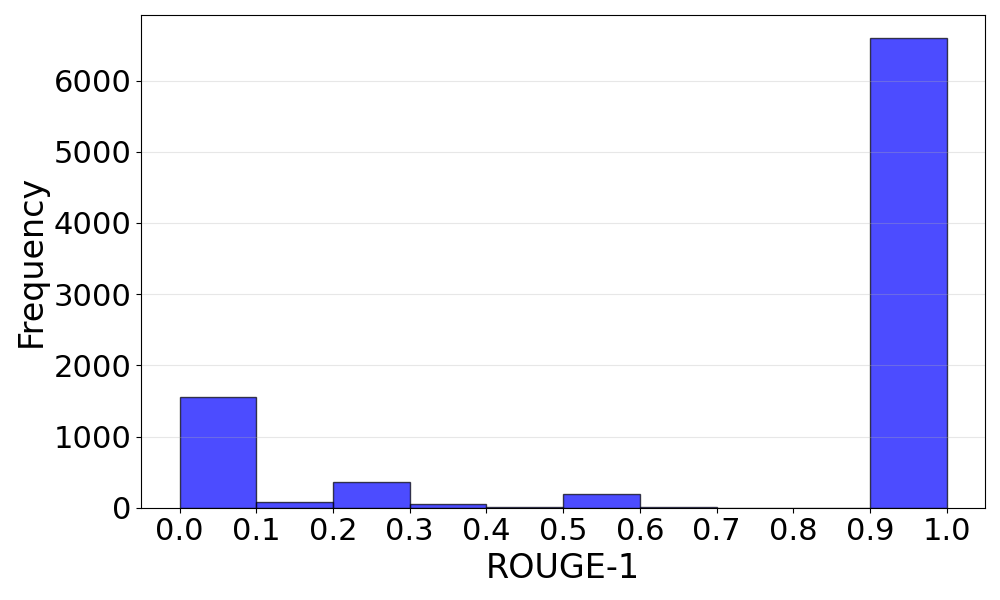}
        \vspace{10pt}
        \caption{\scriptsize CrossTask}
    \end{subfigure}\hfill
    \begin{subfigure}{0.32\textwidth}
        \centering
        \includegraphics[width=0.95\linewidth]{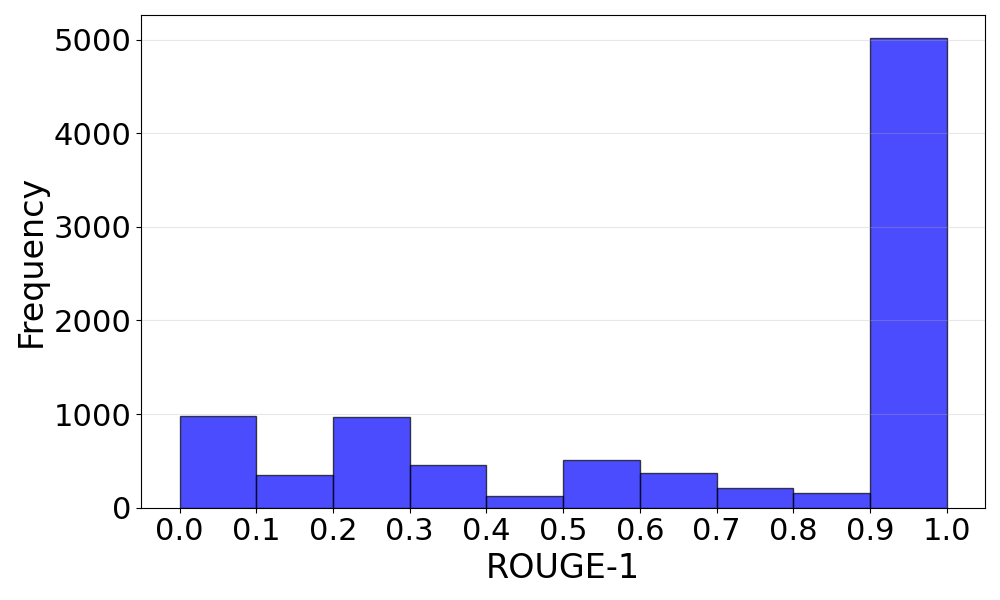}
        \vspace{10pt}
        \caption{\scriptsize Coin}
    \end{subfigure}\hfill
    \begin{subfigure}{0.32\textwidth}
        \centering
        \includegraphics[width=0.95\linewidth]{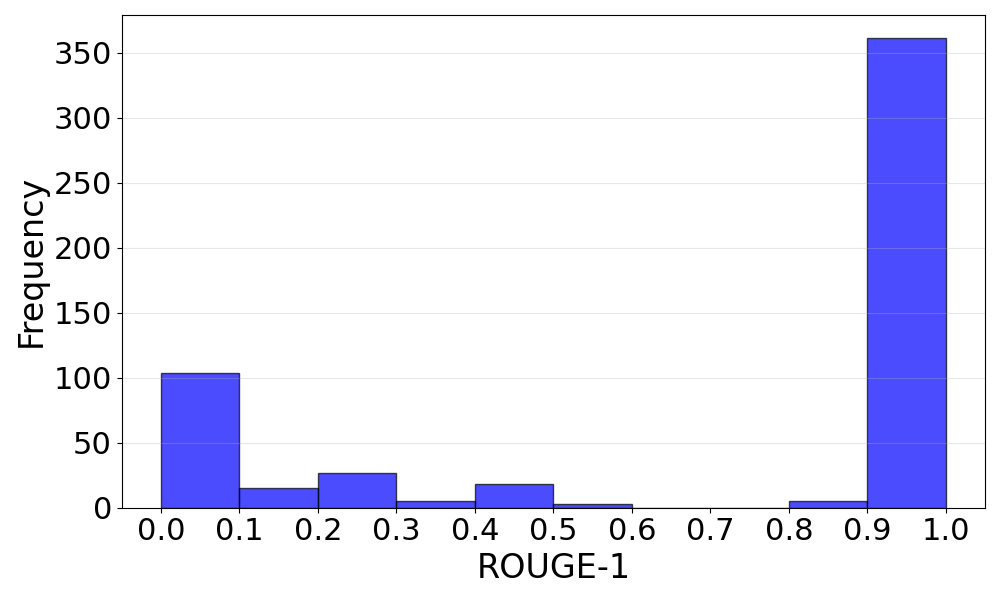}
        \vspace{10pt}
        \caption{\scriptsize NIV}
    \end{subfigure}

    \caption{ROUGE-1 score distribution of \method on CrossTask, Coin and NIV dataset.}
    \label{fig:rouge_dist}
\end{figure*}

Our method \method employs a ROUGE-1 score threshold to determine if a generated description accurately represents a video action. A higher threshold imposes a stricter condition for lexical overlap with the ground truth, i.e. more matching words between the generated description and the ground truth.
\autoref{fig:rouge_dist} shows the distribution of ROUGE-1 scores for generated descriptions on the CrossTask, COIN, and NIV datasets. We observe that the score distributions are dataset-dependent. For CrossTask and NIV, the majority of descriptions with a score above 0.5 are near-perfect matches (scores close to 1.0), suggesting that threshold increases beyond 0.5 have a limited effect. In contrast, the COIN dataset exhibits a more continuous distribution of scores between 0.5 and 0.9, indicating that the chosen threshold value has a greater influence on its results.

\begin{table*}[h]
    \centering
    \caption{Results of using a different threshold for \method on all datasets with time horizon $T=3$. \textbf{Bold} numbers indicate the best results.}
    \label{tab:ablation_thres09}
    \begin{tabular}{lcccc}
        \toprule
        & & SR$\uparrow$ & mAcc $\uparrow$ & mSIoU $\uparrow$ \\
        \midrule
        CrossTask &  KEPP \cite{nagasinghe2024not} (CVPR 2024)	&33.27	&60.16 &/\\
         & \method (TH=0.5) & \textbf{41.14} & \textbf{70.13} & \textbf{65.98} \\
        &\method (TH=0.9) & 37.01	&67.14	&63.02 \\        
        & \method-Text & 35.84 & 65.65 & 62.48\\
        \midrule
        Coin & PlanLLM \cite{yang2025planllm} (AAAI 2025)	&33.22 	&54.33	&/ \\
        &\method (TH=0.5) & \textbf{44.43} & \textbf{68.08} & \textbf{66.71} \\
        & \method (TH=0.9) & 32.58	&55.23	&55.3 \\
        & \method-Text & 27.06 & 50.06 & 52.43 \\
        \midrule
        NIV & MTID \cite{zhoumasked25} (ICLR 2025)	&28.52	&44.44	&56.46 \\
        &\method (TH=0.5) & \textbf{56.51} & \textbf{72.86} & \textbf{72.22} \\
        		
        & \method (TH=0.9) & 53.9	&72.12	&70.84 \\
        & \method-Text & 43.87 & 65.55 & 66.16 \\
        \bottomrule
    \end{tabular}
\end{table*}

To quantify this impact, we evaluate our method \method with a high threshold of 0.9, as shown in \autoref{tab:ablation_thres09}. For CrossTask and NIV, performance of SR slightly drops , i.e. SR decreases from 41.14 to 37.01 on CrossTask and from 56.51 to 53.9 on NIV, yet our method still surpasses the best baselines. On COIN, however, the stricter threshold causes a more significant performance degradation, with the Success Rate falling from 44.43 to 32.58. This is consistent with the score distribution in \autoref{fig:rouge_dist}, as the higher threshold excludes a larger proportion of plausible descriptions for COIN.
This analysis shows a trade-off: a threshold that is too high may discard semantically correct but lexically diverse descriptions, while one that is too low risks incorporating noisy, inaccurate predictions. The choice of the threshold value is therefore dataset-specific and application-specific. 

\subsection{Impact of Unknown Label}

To further analyse the impact of the unknown label from the VLM in \method, we replace the unknown label with a random label when the highest ROUGE-1 score of generated descriptions is below the threshold. 
The high level intuition is that the model will have a random guess about the start or goal action instead of ``saying I do not know''.
\begin{table}[h]
    \centering
    \caption{Comparison of ``Unknown'' label and random label when the generated descriptions of start or goal action have lower ROUGE-1 score than the threshold. \textbf{Bold} numbers are the best. $\dagger$ indicates \method with random label.}
    \label{tab:ablation_random_vs_unknown}
    \begin{tabular}{l l c c c c c c}
        \toprule
        & & \multicolumn{3}{c}{T=3} & \multicolumn{3}{c}{T=4} \\
        \cmidrule(lr){3-5} \cmidrule(lr){6-8}
         &  & SR$\uparrow$ & mAcc$\uparrow$ & mSIoU$\uparrow$ & SR$\uparrow$ & mAcc$\uparrow$ & mSIoU$\uparrow$ \\
        \midrule
        \multirow{2}{*}{CrossTask} 
        & \method $\dagger$ & 39.13 & 67.73 & 62.26 & 23.63 & 60.81 & 58.65 \\
        & \method    & \textbf{41.14} & \textbf{70.13} & \textbf{65.98} & \textbf{23.91} & \textbf{63.70} & \textbf{62.56} \\
        \midrule
        \multirow{2}{*}{Coin} 
        & \method $\dagger$ & 41.16 & 63.70 & 60.78 & 29.12 & 59.48 & 61.43 \\
        & \method    & \textbf{44.43} & \textbf{68.08} & \textbf{66.71} & \textbf{31.56} & \textbf{63.77} & \textbf{68.24} \\
        \midrule
        \multirow{2}{*}{NIV} 
        & \method $\dagger$ & 46.10 & 65.92 & 64.32 & 36.28 & 63.83 & 60.96 \\
        & \method    & \textbf{56.51} & \textbf{72.86} & \textbf{72.22} & \textbf{43.81} & \textbf{68.03} & \textbf{70.17} \\
        \bottomrule
    \end{tabular}
\end{table}

\autoref{tab:ablation_random_vs_unknown} shows the comparison of using ``Unknown'' and random label. \method using ``Unknown'' label consistently outperforms using random label across all datasets on all metrics.
As mentioned in the main text, the model does not see unknown labels during and the correctly predicted start or goal action could still guide the diffusion model to generate correct action sequences. 
On the other hand, if random labels are used, they have been seen during training. Then in inference, a random guess could result in an irrelevant start or goal action and it will make the diffusion model more difficult to generate action sequences. 
We analyse the samples in test set of NIV where either the predicted start or goal action is wrong or both are wrong. In the cases where the start action prediction is correct but the goal is wrong, using unknown label has 24\% more samples that have the correct action sequence than using random label. For the cases where the start is wrong but the goal is correct, using unknown label has 29\% of samples. When both the start and goal action is wrong, it does not matter whether unknown or random label is used, as none of the sample has correct sequences in both cases.

\subsection{Computational Cost}
\begin{table}[h]
    \centering
    \caption{Number of parameters, inference time and throughput.}
    \label{tab:computation}
    \resizebox{\linewidth}{!}{
    \begin{tabular}{ccccc}
        \toprule
        \multicolumn{2}{c}{Inference Time (Seconds)} & \multicolumn{2}{c}{No. Trainable Parameters (M)} & Throughput (Seconds) \\
        \cmidrule(r){1-2} \cmidrule(r){3-4} 
        VLM & Diffusion & VLM & Diffusion & \\
        1.94 & 0.006 & 497.5 & 40.4 & 2.3 \\
        \bottomrule
    \end{tabular}
    }
\end{table}

\autoref{tab:computation} shows the computational cost of \method.
\method has 497.5M trainable parameters for VLM and 40.4M for diffusion model, while MTID has 1.1B and PlanLLM has 133.1M. The average inference time for VLM and the diffusion model is 1.94s and 0.006s, while the throughput is 2.3s on one A100, which is sufficient for real-world applications.

\end{document}